%% file: main.tex
\documentclass{article} 
\usepackage{iclr2027_conference,times}

\usepackage{caption}
\usepackage{wrapfig}
\usepackage{hyperref}
\usepackage{url}
\usepackage{amsmath,amssymb,amsthm}
\usepackage{graphicx}        
\usepackage{booktabs}        
\usepackage{multirow}        
\usepackage{xcolor}          
\usepackage{colortbl}
\usepackage{microtype}       
\usepackage{xspace}          
\usepackage{cleveref}        
\crefname{figure}{Fig.}{Figs.}
\Crefname{figure}{Fig.}{Figs.}
\crefname{table}{Tab.}{Tabs.}
\Crefname{table}{Tab.}{Tabs.}
\crefname{section}{Sec.}{Secs.}
\Crefname{section}{Sec.}{Secs.}
\crefname{subsection}{Sec.}{Secs.}
\Crefname{subsection}{Sec.}{Secs.}
\crefname{equation}{Eq.}{Eqs.}
\Crefname{equation}{Eq.}{Eqs.}
\usepackage{pifont}          
\usepackage{enumitem}        
\usepackage[normalem]{ulem}  

\newcommand{\circnum}[1]{\raisebox{-1.1pt}{\ding{\numexpr181+#1\relax}}}

\definecolor{citecolor}{RGB}{30, 90, 170}   
\definecolor{linkcolor}{RGB}{170, 40, 40}   
\hypersetup{
  colorlinks = true,
  citecolor  = citecolor,
  linkcolor  = linkcolor,
  urlcolor   = citecolor,
}

\setcitestyle{numbers,square,comma}

\newcommand{\method}{\textbf{\textsc{Baton}}\xspace}          
\newcommand{\harness}{Harness VLA\xspace}
\newcommand{\tsr}{TSR\xspace}
\newcommand{\csr}{CSR\xspace}
\newcommand{\vlaact}{\texttt{VLA\_ACT}\xspace}

\newcommand{\pifive}{$\pi_{0.5}$}
\newcommand{\pirlinf}{$\pi_{RLinf}$\xspace}
\newcommand{\subt}[1]{u_{#1}}                        
\newcommand{\plan}{\mathcal{P}}                      
\newcommand{\rest}[1]{\plan_{>#1}}                   
\newcommand{\lib}{\mathcal{M}}                       

\title{%
  \mbox{%
    \raisebox{-0.32\height}[0pt][0pt]{%
      \includegraphics[height=1.6em]{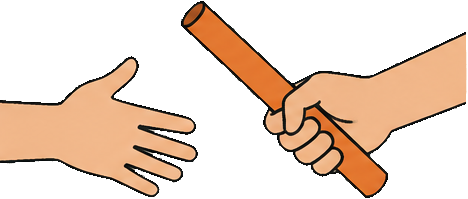}}%
    \hspace{0.3em}Don't Drop the \method:%
  }
  Long-Horizon Robot Manipulation via Agentic Subtask Exploration and Transition-Aware Memory%
}

\author{Bingxin Xu$^{1}$ \quad Yuzhang Shang$^{2}$ \quad Emilio Ferrara$^{1}$ \\
$^{1}$University of Southern California \quad $^{2}$University of Central Florida
}

\iclrfinalcopy 

\begin{document}

\maketitle

\input{abs}
\input{intro}
\input{related}
\input{method}
\input{exp}
\input{con}

\bibliography{references}
\bibliographystyle{iclr2027_conference}





\end{document}

%% file: abs.tex
\begin{abstract}
Long-horizon robot manipulation chains many contact-rich skills into
one multi-stage task. Vision-language-action (VLA) and world-action models (WAMs)
increasingly master the individual skills. However, the chain itself still fails: errors compound beyond the policy's ability to correct, and the execution of one subtask silently constrains the next.
Coding agents offer a promising pathway: freeze the VLA and put an LLM agent in charge of it.
The agent plans in language, moves in free space with analytic primitives, and invokes the VLA only for the contact-rich segments,
writing all adaptation into language memory.
Yet applied directly to long horizons, the agent-plus-VLA recipe breaks twice.
\circnum{1} Its competence is acquired through whole-task exploration
at test time, whose cost is exponential in the number of stages:
where a single stage requires $T$ episodes on average, a $K$-stage
task requires on the order of $T^K$, and a failed episode does not
identify the stage responsible.
\circnum{2} It has no representation of the transitions: the VLA
primitive carries an exit condition but no entry condition, and a
subtask can succeed in a form its successor cannot use.
We present \method, which addresses the two failures in
turn.\footnote{The recipe needs this
interface, not a particular backend: a VLA, a WAM, or any contact-rich policy
callable the same way, slots in unchanged. We write VLA throughout for simplicity.}
Against \circnum{1}, \method makes the subtask the unit of
exploration.
Each subtask is explored in the inexpensive short-horizon regime, and
its solution is stored in memory.
A long-horizon trajectory is then composed from these solutions
rather than discovered whole.
Exploration cost thus becomes linear in the number of stages ($KT$), and every
failure is attributed to a single stage.
Against \circnum{2}, \method equips this exploration with a
transition-aware memory.
Within a subtask, a verifier agent governs the \emph{invocation
transition}: the VLA is invoked only after the wrist view confirms
that the scene is ready.
Across subtasks, \emph{handoff
transition} restores an entry state disturbed by the predecessor's
residue, and \emph{lookahead transition} determines the execution strategy whose
outcome the successor can inherit.
On the long-horizon manipulation benchmark RoboMemArena, \method improves task success rate by 37.7\% and cumulative success rate by 29.7\% over the current SoTA.
\end{abstract}

%% file: intro.tex
\section{Introduction}
\label{sec:intro}

What separates current manipulation policies from generalist ones is less about any single skill than about composition: interpreting an instruction, then holding it together across many contact-rich steps whose errors compound.
The skills themselves are increasingly within reach: vision-language-action
(VLA) models and world-action
models (WAMs) map observations and language directly to low-level actions,
generalizing broadly through web-scale pretraining
\citep{brohan2023rt,kim2024openvla,chi2023diffusionpolicy,black2024pi_0,intelligence2025pi_,bjorck2025groot,shi2025hirobot,liu2026oawam,wang2026worldmodelsurvey}.
What none of them yet delivers is the chain.
Set \pifive{} \citep{intelligence2025pi_} on a single skill (open
a drawer, pick up a can, place it in the microwave) and it executes
cleanly, at $97\%$ success on LIBERO's single-task suites
\citep{openpi2025}; set it on the thousand-step tasks that chain such skills and
it clears $38.7\%$ of the intermediate stages while finishing only
$21.5\%$ of the full tasks \citep{lei2026robomemarena}.
The stages succeed, and the task dies between them.

Two things go wrong.
First, errors compound and the policy cannot repair them. End-to-end VLAs overfit the demonstrated trajectories from their
curated initial states \citep{fang2026liberocf,fei2025liberoplus}.
When each stage starts wherever the previous one ended,
it can only replay memorized motions from states it has never seen,
and it can neither recognize nor correct the accumulating drift
\citep{ross2011dagger,zheng2026failingforward}.
Second, neighboring subtasks are coupled through execution: how one
subtask is carried out fixes the state the next begins from, and an
execution that succeeds locally can leave the next subtask impossible
\citep{lee2021adversarial,yang2025boss,foong2026handful}.

Coding agents are built for exactly this kind of composition
\citep{yang2024sweagent,wang2024voyager,shinn2023reflexion}.
In this line of work \citep{liang2023codeaspolicies,fu2026capx}, an
LLM agent controls the robot by writing programs over perception and
control primitives.
The programs run, and the agent revises them from execution feedback.
Nothing is trained; the agent improves by trying things at test time.
Harness VLA \citep{zhang2026harness} brings this recipe to contact-rich
manipulation by adding a frozen VLA to the agent's toolbox.
The agent plans in language and moves through free space with analytic,
inverse-kinematics-solved primitives; where the task turns
contact-rich, it hands over to the VLA, wrapped as a retryable
primitive with a prompt and a stop predicate.
For long-horizon control, this division of labor is attractive on
three counts.
First, each part does what it is best at: the VLA handles local
contact, and the agent handles composition, semantic re-binding, and
staging.
Second, nothing is fine-tuned. One frozen policy serves every task.
Third, everything the system learns is written in language, so it can
be read, audited, and reused.
On short-horizon tasks, this design already works: skills stay
reliable under perturbations that break end-to-end VLAs
\citep{zhang2026harness}.
Long-horizon manipulation is another matter: no coding agent yet earns
the same competence over a thousand-step chain.

\begin{figure}[t]
  \centering
  \includegraphics[width=0.99\textwidth]{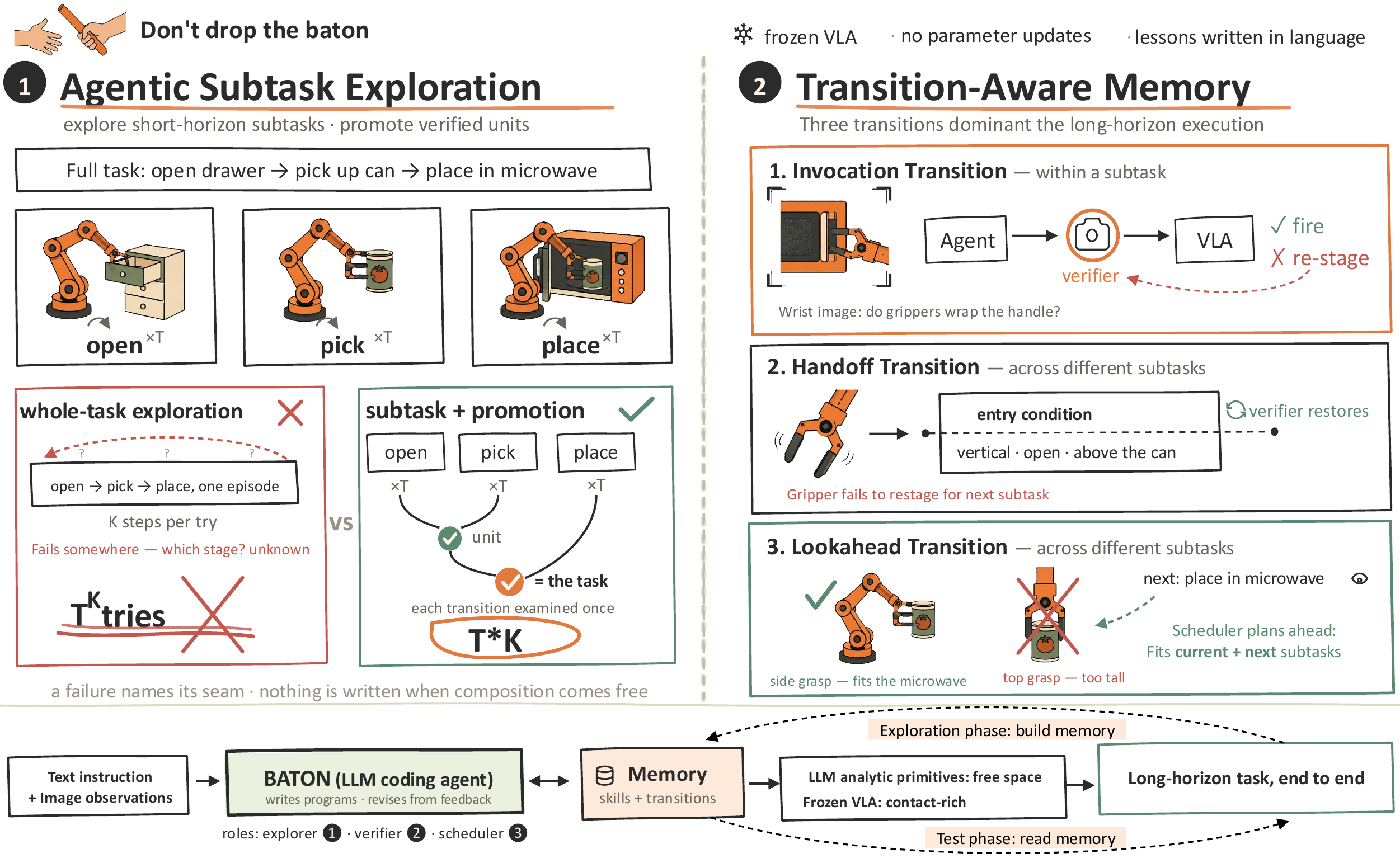}
  \vspace{-0.1in}
  \caption{\textbf{\method{} treats transitions as first-class objects.}
  \emph{Bottom}: overall workflow (\cref{sec:prelim}).
  \emph{Left}: subtask exploration makes the cost linear in the number of stages and localizes
  each failure (\cref{sec:agent}).
  \emph{Right}: the three proposed transition memories (\cref{sec:memory}).}
  \vspace{-0.2in}
  \label{fig:teaser}
\end{figure}

Two difficulties block the way.
\uline{\circnum{1} is the large cost of full task exploration.}
The agent earns its competence at test time. During exploration, the agent freely trials staging orders, pre-contact poses, VLA
invocation timings, and termination thresholds until the task first
succeeds, then consolidates the working solution into memory
\citep{zhang2026harness}.
For a short-horizon task of a hundred-odd steps, first success arrives
after $T$ exploration episodes on average, with a matching bill of
planner tokens: a real price, but one paid once per task.
Stretching the same procedure over a task of $K$ such stages breaks it.
An episode succeeds only if every stage does, so whole-task exploration
needs on the order of $T^K$ episodes, beyond any budget (see \cref{fig:teaser}, left).
To make matters worse, the failures are uninformative: an episode reveals only that the
task failed, not which stage broke.
\uline{\circnum{2} is the transitions the chain must cross.}
Beyond its cost, exploration is uninformed precisely where a chain
is fragile: at the crossings.
Within a subtask, no signal indicates when the scene is ready for
the contact-rich policy, so the moment of handover must be found by
repeated trial.
Across subtasks, individually successful skills fail to compose, for
two reasons.
First, a completed subtask alters the state its successor starts
from. A subsequent failure is therefore ambiguous: the successor's skill
may be inadequate, or its entry state is merely wrong, however exploration
cannot distinguish the two.
Second, even a correct entry is not sufficient: subtasks are
interdependent, and the requirements of the next constrain how the
previous one should be executed.
This dependency spans subtasks, so a learner that acquires them
one at a time can neither observe it nor correct for it.

Targeting \circnum{1}, we propose \method{}, an LLM agent orchestrating a frozen VLA for long-horizon robot manipulation tasks.
\method makes the subtask the unit of exploration: it decomposes a
long-horizon task and explores each subtask in the cheap short-horizon
regime where the recipe can work.
Each solved subtask's successful execution is extracted into abstract memory.
A long-horizon trajectory is then composed from these pieces by memory
rather than discovered whole.
Exploration cost now scales linearly in the number of new subtasks
and boundaries rather than exponentially (see \cref{fig:teaser}, left).
A failure names the stage that caused it instead of voiding a
thousand-step episode.
A relay team does not train by running ten thousand full races; it
drills each exchange.

\method meets \circnum{2} with a
\emph{transition-aware} memory that consists of three parts.
\emph{Within} a subtask lies the \emph{invocation transition}, where
control passes from the agent to the contact-rich VLA after a \emph{verifier agent} confirms.
\emph{Across} subtasks, two transitions govern each boundary.
The \emph{handoff transition} concerns the entry state of the
successor: the boundary records an entry condition, and the verifier
restores the scene to satisfy it before the next subtask begins.
The \emph{lookahead transition} operates in the reverse direction:
the subtask requirements that follows determine how the
current one is executed.
A \emph{scheduler agent} enforces this dependency, selecting from
memory the strategy whose outcome the successor can inherit.
Together these conditions turn each boundary into a \emph{handoff
contract} that is drafted in language, checked at execution, and
stored with the coarse trajectories it connects.
A good decomposition cuts a long task at interfaces that can be checked.
Our core contributions are:
\begin{itemize}[leftmargin=1.4em, itemsep=1pt, topsep=2pt, parsep=0pt]
\item We present \method, the first coding agent to
carry long-horizon robot manipulation end to end over a frozen VLA,
learning by test-time exploration with no parameter updates.
\item \method rests on two mechanisms: \emph{hierarchical subtask
exploration} for \circnum{1}, which keeps exploration cost linear in the number of stages;
\emph{transition-aware memory} for \circnum{2}, which attaches
checkable conditions to the invocation, handoff, and lookahead
transitions.
\item On RoboMemArena's thousand-step household tasks, \method
improves task success by 37.7\% and cumulative success by
29.7\% over the current SoTA.
\end{itemize}

%% file: related.tex
\section{Related Work}
\label{sec:related}


\paragraph{Long-horizon robot manipulation.}
The dominant response to long horizons is to train for them.
Web-scale VLA pretraining keeps improving the individual skills
\citep{brohan2023rt,kim2024openvla,black2024pi_0,intelligence2025pi_},
but the resulting policies are reactive: they predict the next action
from the current observation, overfit the curated initial states of
their demonstrations \citep{fang2026liberocf,fei2025liberoplus}, and
cannot correct their own compounding drift
\citep{ross2011dagger,zheng2026failingforward}.
Memory-augmented VLAs train history into the policy
\citep{shi2026memoryvla,sridhar2025memer}, and memory-oriented
benchmarks measure what remains: on RoboMemArena's thousand-step
household tasks, every reported policy completes far more stages than
full tasks \citep{lei2026robomemarena,dai2026robomme}.
That skills fail when chained where each succeeds alone is
documented across settings.
Sequenced policies need terminal states aligned with what the next
policy expects \citep{lee2021adversarial}; on BOSS, observation-space
shift at subtask boundaries alone cuts the success of four baseline
policies by $34$--$67\%$, a drop that thirty times more demonstrations
do not remove \citep{yang2025boss}; VLA skills with isolated success rates of
$77$--$100\%$ still collapse under composition
\citep{rui2026semantichandoff}; and nine grasping policies with
near-identical first-subtask success (about $95\%$) spread from $0\%$
to $78\%$ on the subtask that follows \citep{foong2026handful}.
The fixes, again, train for the boundary: HANDFUL selects among
finger configurations by curriculum-based successive elimination
\citep{foong2026handful}, and Foresight Residual RL shapes terminal
states with a residual policy rewarded by a learned
downstream-success predictor \citep{liu2026foresight}.
Both bind the lesson to a fixed successor policy in a known sequence
and pay a per-boundary training cost, so every new pairing of skills
pays again.
\method encodes the same insight symbolically, in the manner of a
coding agent.
The handoff contract is drafted in language at plan time, checked at
the boundary, and corrected when it fails.
Because it attaches to the transition rather than to the task, it is
paid for once and reused by every task that crosses the same boundary.

\paragraph{Memory for manipulation agents.}
Skill libraries since Voyager converge to one program per skill name
\citep{wang2024voyager}.
Embodied-memory systems for manipulation key retrieval on the current
instruction, recent observations, or execution history
\citep{torne2026mem,ji2026hime}.
Memory-augmented VLAs build episodic context into the policy itself
\citep{shi2026memoryvla} or retrieve past keyframes from an experience
buffer \citep{sridhar2025memer}; in either case the memory records
what was seen, not which strategy was chosen.
HiMe is the sharpest foil.
Its hierarchy indexes control frequency, it deletes conflicting
entries to keep memory consistent, and its subtask monitor records
\emph{that} a grasp succeeded rather than \emph{which} grasp
\citep{ji2026hime}.
The rival strategies we preserve are, in that design, a defect to
remove; yet which grasp was taken is exactly what the next subtask
depends on.
Task-oriented grasping selects a grasp for a named use, classically
from a fixed semantic taxonomy \citep{murali2020taskgrasp} and more
recently with open-vocabulary use descriptions grounded by foundation
models \citep{tang2025foundationgrasp}; either way, the condition names a use for one object, not the plan that remains.
Task-and-motion planning derives grasp constraints by reasoning
backward over symbolic operators \citep{kaelbling2011hierarchical},
which requires engineered predicates and models
\citep{garrett2021integrated}.
To our knowledge, \method's memory is the first to keep rival
strategies per subtask and to key retrieval on the remaining plan,
through a scheduler that decides \emph{how} each subtask is executed.
The handoff contract, drafted and refined in language without
training, plays the role of these taxonomies, predicates, and value
functions.

\paragraph{Agentic control of frozen VLAs.}
An LLM agent can control a robot by writing programs over perception
and control primitives, revising them from execution feedback with
nothing trained \citep{liang2023codeaspolicies,fu2026capx}; \harness
extends this recipe to contact-rich manipulation by wrapping a frozen
VLA as a retryable primitive \citep{zhang2026harness}
(\cref{sec:prelim}).
ASPIRE runs agentic skill discovery in code, with no learned policy in
the loop \citep{lu2026aspire}.
Adjacent lines place a reasoning agent above the motor level without
writing programs.
VLM planners orchestrate VLAs as callable tools
\citep{lei2026vlastools,chen2026volo}; dual-system models couple a
slow reasoner to a fast policy
\citep{shi2025hirobot,intelligence2025pi_,peng2026cortex}; world-action
models extend the motor level with predictive world modeling
\citep{liu2026oawam,wang2026worldmodelsurvey}.
All of these systems make individual skill invocations reliable, or
discover new skills.
None constrains what one invocation must leave behind for the next,
and none addresses how exploration cost scales once a task spans many
stages.
\method keeps the substrate of \harness and treats the transition
itself as the unit of both exploration and learning.

%% file: method.tex
\section{Method}
\label{sec:method}

\subsection{Preliminaries}
\label{sec:prelim}

We first review the paradigm of coding agents for robot manipulation
\citep{liang2023codeaspolicies,fu2026capx,zhang2026harness}, in which an LLM agent
solves a task by writing programs and delegates only contact-rich motion to a learned
policy.
Within this paradigm, the agent acts through two kinds of primitive.
Analytic primitives execute non-contact motions, such as transporting the
end-effector, reorienting the wrist, and actuating the gripper, and are composed over
symbolic perception queries.
Contact-rich motion is instead handled by a single learned primitive,
$\texttt{VLA\_ACT}(\ell,\varphi)$, which executes a frozen VLA under instruction $\ell$
until the stop predicate $\varphi$ is satisfied.
Accordingly, each skill factors into two segments,
\begin{equation}
u
\;=\;
\underbrace{a}_{\text{non-contact: LLM agent}}
\;\triangleright\;
\underbrace{v}_{\text{contact-rich: } \vlaact},
\label{eq:harness}
\end{equation}
where $a$ denotes the non-contact prefix staged by the agent and $v$ denotes the
contact-rich segment executed by \vlaact{}.
Since the VLA is invoked only from a pre-contact pose staged by the agent, it operates
near its training distribution.
Moreover, no component of the system is trained, and the agent acquires its competence
through test-time exploration, consolidating the first successful solution into a
language memory $\lib$ that is consulted in subsequent attempts.
It is worth noting that \cref{eq:harness} does not explicitly represent the transition
between $a$ and $v$, and the moment at which the VLA should be invoked is therefore left
to be discovered by trial.

\textbf{Challenges of long-horizon tasks.}
Although this paradigm is effective for short-horizon tasks, where test-time
exploration is affordable and incurs only a one-time cost, it does not extend directly
to long-horizon tasks.
We define a long-horizon task $\mathcal{T}$ as an instruction whose execution
decomposes into an ordered sequence of subtasks $\langle \subt{1}, \dots, \subt{K}
\rangle$, each of which is a skill of the form in \cref{eq:harness}.
Compared with a single skill, such a task exhibits two distinctive properties (see \cref{fig:teaser}).
The first is conjunctive success, as $\mathcal{T}$ succeeds only if every subtask
$\subt{k}$ succeeds.
The second is the interdependence among subtasks.
Since the initial state of $\subt{k+1}$ is determined by the terminal state of
$\subt{k}$, the execution of each subtask determines the starting condition of the next,
while the requirements of the next subtask in turn constrain how the preceding one should
be executed.
As discussed below, these two properties give rise to two corresponding obstacles.

\textbf{Obstacle \circnum{1}: the cost of exploration.}
\label{sec:obstacle}
The first obstacle follows from conjunctive success, which requires whole-task
exploration to find a working realization of every subtask within a single episode.
If a single subtask reaches its first success after $T$ episodes on average, then $K$
subtasks explored jointly require on the order of $T^K$ episodes, since each episode
must succeed on all of them.
Moreover, failures are uninformative.
An unsuccessful episode reveals only that the chain broke, and because the
paradigm consolidates only whole-task solutions, the subtasks solved along the way leave
nothing in memory.
Consequently, exploration that is affordable and self-correcting on a single subtask
becomes prohibitively expensive over a chain and retains no partial progress.

\textbf{Obstacle \circnum{2}: the fragility of transitions.}
The second obstacle follows from the interdependence among subtasks.
As noted above, the formulation in \cref{eq:harness} specifies what each segment should
accomplish but imposes no condition on the transition from $a$ to $v$, and the same
holds for the boundary between consecutive subtasks.
Over a long chain, these unconstrained transitions become the points at which execution
breaks down.
Within a subtask, the paradigm has no notion of when the scene is ready for the VLA, so
the handover is tied to the states encountered during exploration.
Between subtasks, the missing conditions matter in both directions.
In the forward direction, a subtask inherits the residue of its predecessor.
A subsequent failure then becomes ambiguous, since it may stem from either the skill
itself or its entry state.
In the backward direction, a local success admits many realizations, not all of which
the successor can build upon.
However, the paradigm stores only the first solution that succeeds, without recording
the conditions under which it holds.
Whether consecutive subtasks fit together is therefore never checked.

Together, these two obstacles lead to \method's design.
To address the first, exploration must be organized per subtask, so that its cost grows linearly with the chain, each failure is attributed to the subtask responsible, and solved subtasks are retained (\cref{sec:agent}).
To address the second, the transitions both within and between subtasks must be stated as explicit conditions that the system verifies and remembers, rather than left implicit in whichever solution succeeds first (\cref{sec:memory}).

\begin{figure}[t]
  \centering
  \includegraphics[width=0.99\textwidth]{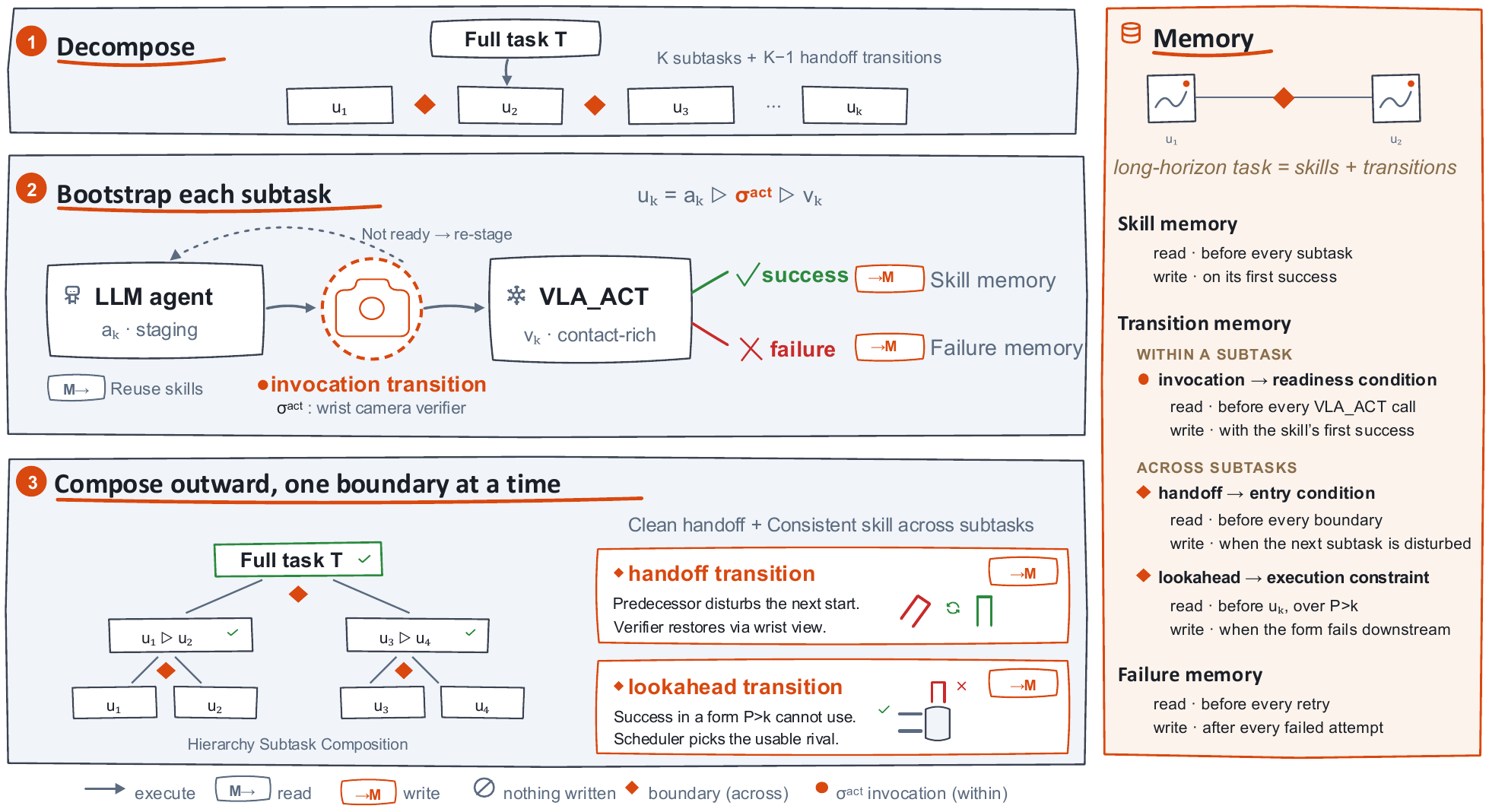}
  \vspace{-0.1in}
  \caption{\textbf{Overview of \method{}.}
  \emph{Left}: \method{} decomposes a long-horizon task into subtasks, explores each
  subtask individually (\cref{sec:agent}), and composes them into the full task through
  three transition memories (\cref{sec:memory}).
  \emph{Right}: the skill, transition, and failure memories built during exploration.}
  \vspace{-0.2in}
  \label{fig:overview}
\end{figure}

\subsection{Hierarchical Subtask Exploration}
\label{sec:agent}

Rather than exploring a long-horizon task as a whole, \method{} represents it as a chain
of subtasks joined by handoff boundaries,
\begin{equation}
\mathcal{T}
\;=\;
\subt{1} \,\triangleright\, \sigma_{1,2} \,\triangleright\,
\subt{2} \,\triangleright\, \sigma_{2,3} \,\triangleright\,
\cdots \,\triangleright\, \sigma_{K-1,K} \,\triangleright\, \subt{K},
\label{eq:chain}
\end{equation}
where each $\subt{k}$ is a short-horizon skill of the form in \cref{eq:harness} and
$\sigma_{k,k+1}$ is the boundary between $\subt{k}$ and $\subt{k+1}$.
We refer to the ordered subtasks $\plan = \langle \subt{1}, \dots, \subt{K} \rangle$ as
the plan.
Exploration proceeds bottom-up over this chain.
Each subtask is first explored in isolation, which keeps the search within the
short-horizon regime where the paradigm of \cref{sec:prelim} is effective.
The agent refines its solution from execution feedback until the subtask first succeeds.
Each failed attempt is recorded in memory, so that later attempts avoid repeating it.
The successful solution is stored in an abstract, symbolic form, so that it can be
reused across instances and tasks.
The solved subtasks are then composed hierarchically.
Adjacent subtasks are chained into pairs, and each successful pair becomes a verified
unit.
These units are in turn chained with their neighbors until the full task runs end to
end.
When a composition fails, both of its parts are already known to succeed on their own,
so the search is directed to the boundary between them.
A boundary is regarded as crossed once the successor succeeds from the state left by
its predecessor.

Compared with whole-task exploration, this bottom-up scheme resolves the first obstacle
in three respects.
First, each round of search targets a single new element, either a subtask or a
boundary between two verified units.
Each element is therefore explored only once, and the total cost scales on the order of
$KT$ rather than $T^K$.
Second, a failure is confined to the element under exploration and never invalidates a
verified unit.
Solved subtasks and compositions are thus retained as reusable progress.
Third, composition introduces no new skills, since each unit simply executes its parts
in order.
Memory therefore grows only at boundaries where composition fails.
These boundaries correspond to the transitions identified in the second obstacle, and
they are the focus of \cref{sec:memory}.

\subsection{Transition-Aware Memory}
\label{sec:memory}

Hierarchical exploration (\cref{sec:agent}) makes the subtasks and their boundaries
explicit, but it leaves the transitions within and between subtasks unconstrained.
\method{} addresses this obstacle with a transition-aware memory $\lib$.
We identify three transitions, each with its own entry in $\lib$.
One lies within a subtask.
The \emph{invocation transition} governs when control passes from the agent to the VLA.
The other two lie at the boundary $\sigma_{k,k+1}$ between consecutive subtasks.
The \emph{handoff transition} governs the state that a subtask delivers to its
successor, while the \emph{lookahead transition} governs how the requirements of the
successor shape the execution of the current subtask.
We describe each transition in turn, and then present the unified store that organizes
them and shares them across tasks.

\textbf{Within a subtask: the invocation transition.}
To represent the transition left implicit in \cref{eq:harness}, \method{} makes it
explicit in the factorization of each subtask,
\begin{equation}
\subt{k}
\;=\;
\underbrace{a_k}_{\text{non-contact: LLM agent}}
\,\triangleright\;
\sigma^{\mathrm{act}}_k
\;\triangleright\,
\underbrace{v_k}_{\text{contact-rich: } \vlaact},
\label{eq:subtask}
\end{equation}
where $\sigma^{\mathrm{act}}_k$ denotes the invocation transition at which control
passes from the agent to the VLA.
Rather than searching for when to invoke the VLA, \method{} conditions this transition
on a \emph{readiness condition} stated in language.
Before each invocation, an agent termed verifier checks the condition
against the wrist-camera observation, and the agent re-stages the scene whenever it
does not hold.
When placing an object into a basket, for instance, the condition may require the
gripper to lie within the rim of the basket.
The readiness condition serves two purposes.
During exploration, it guides the search directly, replacing trial over invocation
timings.
During execution, it keeps the VLA from being invoked in an unsuitable state.

\textbf{Across subtasks (i): the handoff transition.}
The handoff transition addresses the forward direction, in which a predecessor
disturbs the entry state of its successor, such as a gripper left rotated after opening
a door.
Since neither skill is at fault in this case, re-exploring either one would not resolve
the failure.
The boundary $\sigma_{k,k+1}$ instead records an \emph{entry condition}, which specifies
what the scene must satisfy for $\subt{k+1}$ to begin.
This condition belongs to the boundary rather than to either subtask, since the residue
that arrives depends on the predecessor while the residue that matters depends on the
successor.
The verifier checks the condition at the boundary and, when it is violated, restores the
scene with analytic primitives before the successor begins.
Unlike the initiation sets of classical skill formalisms
\citep{sutton1999options,konidaris2009skill}, which are learned offline for each skill,
the entry condition is expressed in language and verified on live observations.

\textbf{Across subtasks (ii): the lookahead transition.}
The lookahead transition addresses the backward direction, in which a subtask succeeds
in a form that its successor cannot use.
An object may be grasped in several ways, for instance, but only some of them leave
enough clearance for placing it into a confined space.
Which execution is appropriate therefore depends not on the subtask alone but on what
follows it.
The boundary $\sigma_{k,k+1}$ accordingly records an \emph{execution constraint}, which
specifies the form in which $\subt{k}$ must end for $\subt{k+1}$ to proceed.
Before executing $\subt{k}$, \method{} looks ahead to the remaining plan
$\rest{k} = \langle \subt{k+1}, \dots, \subt{K} \rangle$ and carries out $\subt{k}$ in a
form that satisfies the constraints it induces.
The same subtask is thus executed differently depending on its successor.

In summary, hierarchical subtask exploration acquires a long-horizon solution piece by
piece, confining each search to a single subtask or boundary.
The transition-aware memory complements it at the transitions that prove fragile,
recording the conditions under which adjacent pieces fit together and verifying them
during execution.
By keeping exploration cost linear in stage count and making fragile transitions explicit, \method{}
carries a long-horizon task reliably from one subtask to the next.

%% file: exp.tex
\section{Experiments}
\label{sec:exp}

\subsection{Experimental Setup}
\label{sec:setup}

\paragraph{Benchmark.}
RoboMemArena \citep{lei2026robomemarena} consists of 26 household
tasks in partially observable kitchen scenes, each chaining $3$--$9$
verification stages over more than $1{,}000$ control steps on
average.
The tasks fall into four categories by what must be carried across
stages: \emph{transferring} (4 tasks), \emph{occlusion} (11),
\emph{counting} (7), and \emph{sequence} (4).
Each task is instantiated by a scene seed that resamples object
placement.
All exploration runs on a single reference seed and evaluation runs
on ten held-out seeds.
We also evaluate \method on a real robot, a PiPER-X 6-DoF research
arm, on two long-horizon tasks with 20 trials each.
We report two metrics: task success rate (\tsr), the percentage of
episodes that complete the task, and cumulative success rate (\csr),
the percentage of stages completed.
A high \csr with a low \tsr means that the stages succeed and the
chain does not.

\paragraph{Models.}
For a controlled comparison, \method{} and the \harness{} baseline
\citep{zhang2026harness} are built on identical components.
Both use the \pirlinf{} checkpoint \citep{yu2026rlinf} as the frozen VLA policy without
any fine-tuning, SAM-3 \citep{sam3} for segmentation, and GPT-5.6-sol as the backbone of the
coding agent, and both follow the coding-agent loop, analytic primitives, and skill
memory of \harness{}.
Both systems start from the skill memory that \harness{} acquires on LIBERO
\citep{liu2023libero}, whose skills closely resemble the subtasks of RoboMemArena, and
both explore on the reference seed with a budget of 20 episodes per task.
\harness{} spends this budget exploring each task \emph{as a whole}, whereas \method{}
spends it exploring the transitions between subtasks.
Performance differences between the two can therefore be attributed to these components
rather than to the surrounding infrastructure.
We further compare \method{} with previous methods evaluated on RoboMemArena
\citep{lei2026robomemarena}, including a VLM-only agent, the reactive VLA
\pifive{} \citep{intelligence2025pi_}, and memory-augmented VLAs
\citep{shi2026memoryvla,sridhar2025memer,dai2026robomme,lei2026robomemarena,lin2026hif}.

\begin{table}[t]
\caption{RoboMemArena \citep{lei2026robomemarena}: per-category and
average \tsr and \csr, in \%. We compare with previous methods and with
\harness, which shares all components with \method. The ground-truth
oracle executes with perfect memory of the past and bounds what
retrospection alone can achieve; it is excluded from bolding. Best per
column in bold.}
\label{tab:main}
\vspace{-0.1in}
\centering
\small
\renewcommand{\arraystretch}{0.92}
\setlength{\tabcolsep}{3.6pt}
\begin{tabular}{lcccccccccc}
\toprule
& \multicolumn{2}{c}{Transferring} & \multicolumn{2}{c}{Occlusion} & \multicolumn{2}{c}{Counting} & \multicolumn{2}{c}{Sequence} & \multicolumn{2}{c}{Average} \\
\cmidrule(lr){2-3}\cmidrule(lr){4-5}\cmidrule(lr){6-7}\cmidrule(lr){8-9}\cmidrule(lr){10-11}
Method & \tsr & \csr & \tsr & \csr & \tsr & \csr & \tsr & \csr & \tsr & \csr \\
\midrule
GPT-5.4 (VLM only) & 13.8 & 32.9 & 1.8 & 9.2 & 12.9 & 50.7 & 15.0 & 47.3 & 8.7 & 30.5 \\
\pifive{} \citep{intelligence2025pi_} & 20.0 & 42.8 & 12.7 & 17.2 & 14.3 & 50.9 & 60.0 & 71.6 & 21.5 & 38.7 \\
HiF-VLA \citep{lin2026hif} & 17.5 & 38.9 & 12.7 & 27.1 & 8.6 & 45.9 & 42.5 & 70.2 & 16.9 & 39.8 \\
MemoryVLA \citep{shi2026memoryvla} & 15.0 & 37.2 & 7.3 & 13.1 & 14.3 & 55.1 & 37.5 & 65.2 & 15.0 & 35.3 \\
MemER \citep{sridhar2025memer} & 20.0 & 36.1 & 16.4 & 33.2 & 27.1 & 65.1 & 65.0 & 79.1 & 27.3 & 49.1 \\
PrediMem \citep{lei2026robomemarena} & 22.5 & 45.2 & 27.3 & 38.4 & 45.7 & 69.3 & 72.5 & 89.5 & 38.5 & 55.2 \\
\textit{PrediMem + GT memory} \citep{lei2026robomemarena} & \textit{32.5} & \textit{54.8} & \textit{33.6} & \textit{49.8} & \textit{51.4} & \textit{75.6} & \textit{85.0} & \textit{92.3} & \textit{46.1} & \textit{64.8} \\
FrameSamp+Modul \citep{dai2026robomme} & 63.8 & 72.1 & 39.1 & 56.5 & 31.4 & 57.8 & 73.8 & 86.9 & 46.1 & 63.9 \\
\midrule
\harness \citep{zhang2026harness} & 50.0 & 58.3 & 0.0 & 2.3 & 14.3 & 50.8 & 75.0 & 93.1 & 23.1 & 37.9 \\
\rowcolor{blue!15} \method (ours) & \textbf{75.0} & \textbf{83.3} & \textbf{50.0} & \textbf{78.6} & \textbf{64.3} & \textbf{83.3} & \textbf{87.5} & \textbf{93.8} & \textbf{63.5} & \textbf{82.9} \\
\bottomrule
\end{tabular}
\vspace{-0.2in}
\end{table}

\subsection{Main Results}
\label{sec:main}

As reported in \cref{tab:main}, \method{} establishes a new state of the art on
RoboMemArena, attaining a \tsr{} of $63.5\%$ and a \csr{} of $82.9\%$.
These results exceed the strongest prior method by 37.7\% and 29.7\%,
and \harness{} by $40.4$ and $45.0$ points, respectively.
The superiority of \method{} further persists across all task categories, with \tsr{}
gains of $11.2$, $10.9$, $18.6$, and $12.5$ points over the best competing method on
transferring, occlusion, counting, and sequence, respectively.
It is worth emphasizing that these results are achieved with a frozen VLA that receives
no fine-tuning on the benchmark, in contrast to the reported policies, which are trained
on its data.

A closer examination of the results reveals two insights into the source of these gains.
First, \harness{} fails almost entirely on the occlusion category.
Within the 20-episode budget, whole-task exploration rarely reaches a first success on these many-stage tasks and thus
  writes nothing to memory, whereas BATON reaches a first success on every subtask and boundary within the same budget.
More fundamentally, occlusion tasks compose a heterogeneous set of operations, such as
opening and closing drawers and microwaves, and thus exhibit a substantially stronger
dependence on inter-subtask transitions than the remaining categories.
The pronounced improvement of \method{} on this category therefore directly reflects the
contribution of the transition-aware memory.
Second, \harness{} already attains strong performance on the sequence category, in which
subtasks are only loosely coupled and transitions are correspondingly less
consequential.
The comparatively modest gain of \method{} on this category is consistent with this
interpretation.
The benefit of \method{} scales with the
extent to which a task depends on the transitions between its subtasks.

\begin{wraptable}{r}{0.55\textwidth}   
\vspace{-0.2in}
\centering
\small
\renewcommand{\arraystretch}{0.92}
\setlength{\tabcolsep}{3pt}
\caption{Real-robot long-horizon experiment results. 
\tsr / \csr in \%, 20 trials per
task. }
\label{tab:real_exp}
\vspace{-0.1in}
\resizebox{\linewidth}{!}{%
\begin{tabular}{l cc cc cc}
\toprule
& \multicolumn{2}{c}{Cans$\to$plate} & \multicolumn{2}{c}{Can+bottle$\to$plate} & \multicolumn{2}{c}{Average} \\
\cmidrule(lr){2-3}\cmidrule(lr){4-5}\cmidrule(lr){6-7}
Method & \tsr & \csr & \tsr & \csr & \tsr & \csr \\
\midrule
\pifive & 10.0 & 31.7 & 5.0 & 21.7 & 7.5 & 26.7 \\
GPT-5.6-sol (VLM) & 15.0 & 38.3 & 5.0 & 25.0 & 10.0 & 31.7 \\
\harness \citep{zhang2026harness} & 35.0 & 61.7 & 20.0 & 43.3 & 27.5 & 52.5 \\
\rowcolor{blue!15} \method (ours) & \textbf{55.0} & \textbf{73.3} & \textbf{40.0} & \textbf{63.3} & \textbf{47.5} & \textbf{68.3} \\
\bottomrule
\end{tabular}}
\vspace{-0.1in}
\end{wraptable}
As shown in \cref{tab:real_exp}, \method{} remains effective on a physical robot.
We evaluate two long-horizon tasks, namely placing two cans onto a plate in sequence and
placing a can and a bottle onto a plate, and compare \method{} with \pifive{}, a vanilla
coding agent, and \harness{}, where the latter two share the GPT-5.6-sol backbone of
\method{}.
The baselines exhibit distinct failure modes.
\pifive{} fails in a largely consistent manner, as it tends to follow a fixed
trajectory regardless of the state left by the preceding subtask.
The vanilla agent adapts more flexibly to such states, yet lacks the precision required
for contact-rich execution.
\harness{} executes contact precisely whenever the preceding transition is appropriate,
but remains vulnerable to failures at the transitions themselves.
By additionally securing these transitions, \method{} achieves the best performance.

\begin{table}[t]
\caption{Ablation of the three transition memories on RoboMemArena:
per-category and average \tsr and \csr, in \%. Each variant removes one
transition of \cref{sec:memory} and keeps the hierarchical subtask
exploration of \cref{sec:agent}. Best per column in bold.}
\vspace{-0.1in}
\label{tab:ablation}
\centering
\small
\renewcommand{\arraystretch}{0.92}
\setlength{\tabcolsep}{3.6pt}
\begin{tabular}{lcccccccccc}
\toprule
& \multicolumn{2}{c}{Transferring} & \multicolumn{2}{c}{Occlusion} & \multicolumn{2}{c}{Counting} & \multicolumn{2}{c}{Sequence} & \multicolumn{2}{c}{Average} \\
\cmidrule(lr){2-3}\cmidrule(lr){4-5}\cmidrule(lr){6-7}\cmidrule(lr){8-9}\cmidrule(lr){10-11}
Variant & \tsr & \csr & \tsr & \csr & \tsr & \csr & \tsr & \csr & \tsr & \csr \\
\midrule
w/o invocation & 52.5 & 63.5 & 36.4 & 66.5 & 42.9 & 61.8 & 80.0 & 93.4 & 47.3 & 68.9 \\
w/o handoff & 57.5 & 67.1 & 9.1 & 44.4 & 21.4 & 78.5 & 77.5 & 93.3 & 30.4 & 64.6 \\
w/o lookahead & 62.5 & 70.8 & 18.2 & 47.8 & 28.6 & \textbf{83.3} & 82.5 & \textbf{93.8} & 37.7 & 68.0 \\
\rowcolor{blue!15} \method (full) & \textbf{75.0} & \textbf{83.3} & \textbf{50.0} & \textbf{78.6} & \textbf{64.3} & \textbf{83.3} & \textbf{87.5} & \textbf{93.8} & \textbf{63.5} & \textbf{82.9} \\
\bottomrule
\end{tabular}
\vspace{-0.2in}
\end{table}

\subsection{Ablation: Which Transition Matters}
\label{sec:ablation}

To isolate the contribution of each transition memory, we conduct an ablation study on
RoboMemArena.
All variants retain the hierarchical subtask exploration of \cref{sec:agent} and follow
the evaluation protocol of \cref{sec:setup}, while the three transition memories of
\cref{sec:memory} are removed one at a time.
Differences in performance can therefore be attributed to the memory rather than to the
exploration.
In the \emph{w/o invocation} variant, the VLA is invoked without checking the readiness
condition, as in \harness{}.
In \emph{w/o handoff}, the boundary records no entry condition, and
each subtask starts from whatever state its predecessor leaves.
In \emph{w/o lookahead}, the scheduler is disabled, and memory keeps
only the first strategy that succeeded for each subtask, as in prior
skill libraries.

The results in \cref{tab:ablation} yield three key observations.
First, each transition memory is indispensable to \method{}.
Ablating any single transition substantially degrades the overall performance, reducing
\tsr{} and \csr{} by $39.4\%$ and $19.0\%$ on average, respectively.
Among the three, the two inter-subtask transitions incur a markedly larger degradation
than the invocation transition, underscoring that the coupling between consecutive
subtasks constitutes the primary determinant of long-horizon success.
Second, the ablations exert a considerably stronger impact on \tsr{} than on \csr{}.
Given that \tsr{} requires the successful completion of every subtask, whereas \csr{}
credits partial progress, this discrepancy corroborates our central premise that
long-horizon failures stem predominantly from the transitions between subtasks rather
than from the subtasks themselves.
Third, the degradation is most pronounced on the occlusion and counting categories.
Compared with transferring and sequence, tasks in these categories entail more frequent
gripper reconfiguration and more stringent constraints across subtask boundaries,
and thus benefit most substantially from the transition-aware memory.
Transferring is the one category where removing the invocation transition hurts
most, since its grasps on the cabinet top depend almost entirely on the wrist-view
alignment of the readiness condition.

\subsection{Qualitative Analysis}
\label{sec:qual}

\begin{wrapfigure}{r}{0.5\linewidth}
  \centering
  \vspace{-0.13in}
  \includegraphics[width=\linewidth]{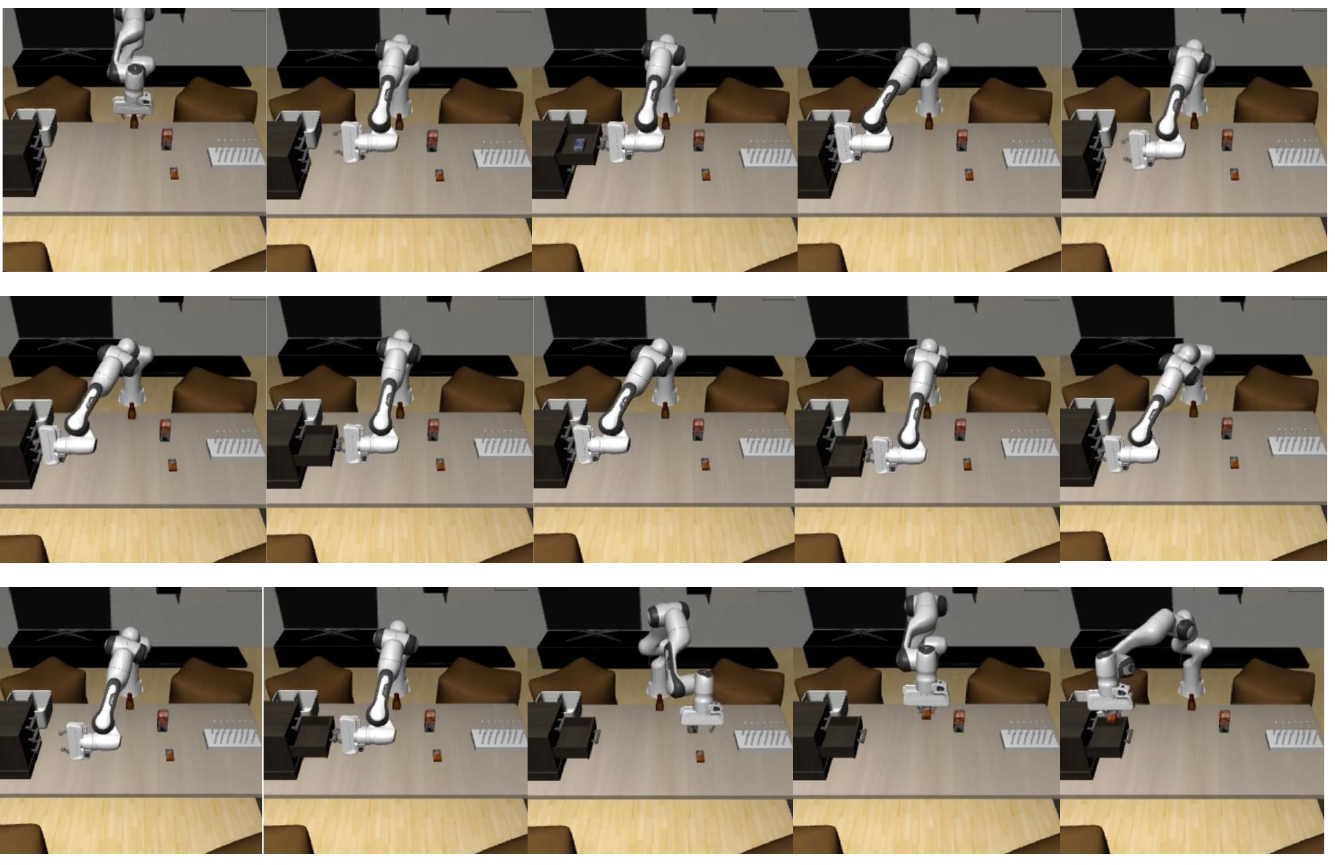}
  \vspace{-0.1in}
  \caption{\textbf{A successful evaluation rollout of \method{} on a
  1,800-step RoboMemArena task.}
  The robot opens and closes three drawers in turn, then places the
  butter in the empty one.}
  \label{fig:demo}
  \vspace{-0.15in}
\end{wrapfigure}
\cref{fig:demo} presents a successful evaluation rollout of \method{} on an occlusion
task under a held-out seed.
The task requires the robot to open and close three drawers in turn, then place the butter in the empty one.
In comparison, \harness{}, which explores the task as a whole, fails to reach a single
success on this task within its exploration budget.
Although it already possesses a skill for opening a drawer, it stalls at this stage,
repeatedly adjusting its approach without securing the handle until the episode is
exhausted.
\method{} avoids this failure by solving each subtask once and composing the solutions,
and the resulting rollout exercises all three transitions of \cref{sec:memory}.
At each drawer, the invocation transition defers the VLA until the wrist view shows the
handle between the fingers.
At each boundary, the handoff transition restores the wrist from the rotated pose left
by the previous opening to the pose expected by the next reach.
At the first drawer, the lookahead transition selects a grasp that keeps the handle
reachable for the subsequent closing motion.
Each transition thus intervenes precisely where the chain requires it.

\begin{wrapfigure}{r}{0.5\textwidth}
    \centering
    \vspace{-0.15in}
    \includegraphics[width=\linewidth]{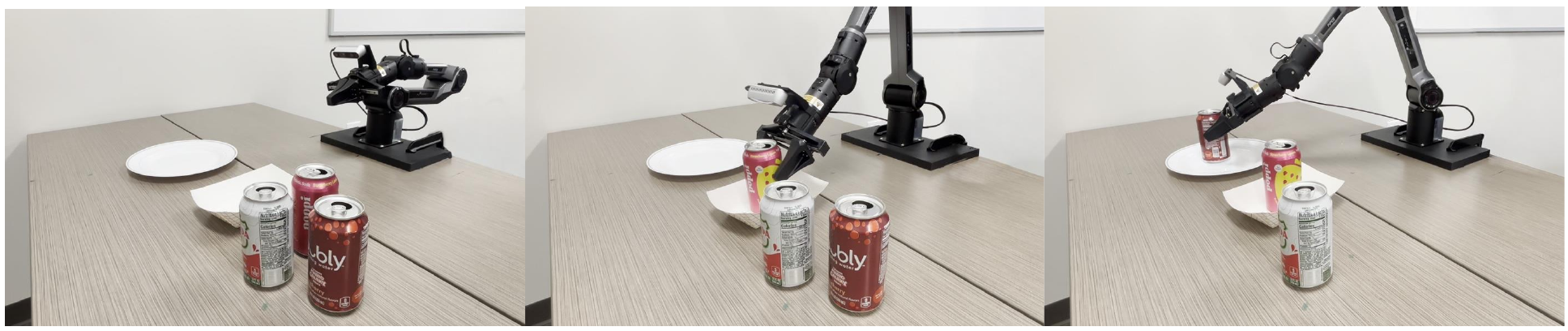}
    \vspace{-0.2in}
    \caption{Real-robot rollout of \method{}. The arm transfers two
    cans onto the plate in turn.}
    \label{fig:real_robot}
    \vspace{-0.15in}
\end{wrapfigure}
\method{} also transfers to the real robot, as illustrated by the successful
long-horizon rollout in \cref{fig:real_robot}.
The real-world setting exhibits the same challenge observed in simulation.
Chaining subtasks is straightforward when they leave the gripper configuration
unchanged.
When a subtask instead requires the gripper to be reoriented or realigned with an
object, the state left by its predecessor no longer matches what it expects, and
chaining fails without transition memory.
\method{} resolves such cases at the transitions, restoring the entry state before each
subtask begins and invoking the VLA only once the scene is ready.

%% file: con.tex
\section{Conclusion}
\label{sec:conclusion}

Long-horizon manipulation is limited less by individual skills than by their composition, since stages that succeed in isolation still fail once chained.
Coding agents over a frozen VLA can compose skills without training, yet none has carried a thousand-step chain, since whole-task exploration is exponentially costly and subtask transitions go unrepresented.
We presented \method{}, to our knowledge the first coding agent to accomplish long-horizon manipulation over a frozen VLA with no parameter updates.
It explores one subtask at a time, so that exploration cost grows linearly with the number of stages, and it states the invocation, handoff, and lookahead transitions as conditions verified during execution.
On RoboMemArena, \method{} improves task success by 37.7\% and cumulative success by 29.7\% over the prior state of the art, with the largest gains where tasks depend most on transitions.
More broadly, our results suggest that the bottleneck of long-horizon manipulation lies in the transitions rather than the skills, and that a policy is made reliable over long horizons not by retraining but by governing when it acts and what it inherits.


